\documentclass{article}

\usepackage{microtype}
\usepackage{graphicx}
\usepackage{subfigure}
\usepackage{booktabs} % for professional tables
\usepackage{mathtools}
\usepackage[table]{xcolor}
\usepackage[normalem]{ulem}

\usepackage{amsmath}
\usepackage{amsfonts}

\newcommand{\tred}[1]{\textcolor{red}{#1}}
\DeclarePairedDelimiter\ceil{\lceil}{\rceil}
\definecolor{lightgray}{gray}{0.9}

\usepackage{hyperref}

\newcommand{\theHalgorithm}{\arabic{algorithm}}

\usepackage[accepted]{icml2021}

\icmltitlerunning{GradFreeBits: Gradient Free Bit Allocation for Dynamic Low Precision Neural Networks}

\begin{document}

\twocolumn[
\icmltitle{
GradFreeBits: Gradient Free Bit Allocation for Dynamic Low Precision Neural Networks}

% It is OKAY to include author information, even for blind
% submissions: the style file will automatically remove it for you
% unless you've provided the [accepted] option to the icml2020
% package.

% List of affiliations: The first argument should be a (short)
% identifier you will use later to specify author affiliations
% Academic affiliations should list Department, University, City, Region, Country
% Industry affiliations should list Company, City, Region, Country

% You can specify symbols, otherwise they are numbered in order.
% Ideally, you should not use this facility. Affiliations will be numbered
% in order of appearance and this is the preferred way.
\icmlsetsymbol{equal}{*}

\begin{icmlauthorlist}
\icmlauthor{Benjamin Bodner}{BU}
\icmlauthor{Gil Ben Shalom}{BGU}
\icmlauthor{Eran Treister}{BGU}
\end{icmlauthorlist}

\icmlaffiliation{BU}{Department of Physics, Brown University, Providence, RI, USA.}
\icmlaffiliation{BGU}{Ben-Gurion University, Beer Sheva, Israel.}

\icmlcorrespondingauthor{Benjamin Bodner}{{\tt benjybo7@gmail.com}}
% \icmlcorrespondingauthor{Eee Pppp}{ep@eden.co.uk}

% You may provide any keywords that you
% find helpful for describing your paper; these are used to populate
% the "keywords" metadata in the PDF but will not be shown in the document
\icmlkeywords{Machine Learning, ICML, Quantization, Gradient Free Optimization, Low Precision Neural Networks, Compression}

\vskip 0.3in
]

% this must go after the closing bracket ] following \twocolumn[ ...

% This command actually creates the footnote in the first column
% listing the affiliations and the copyright notice.
% The command takes one argument, which is text to display at the start of the footnote.
% The \icmlEqualContribution command is standard text for equal contribution.
% Remove it (just {}) if you do not need this facility.

\printAffiliationsAndNotice{}  % leave blank if no need to mention equal contribution
%\printAffiliationsAndNotice{\icmlEqualContribution} % otherwise use the standard text.
% \printNotice{}

\begin{abstract}
% TODO reduce to 4-6 sentences
Quantized neural networks (QNNs) are among the main approaches for deploying deep neural networks on low resource edge devices. Training QNNs using different levels of precision throughout the network (dynamic quantization) typically achieves superior trade-offs between performance and computational load. However, optimizing the different precision levels of QNNs can be complicated, as the values of the bit allocations are discrete and difficult to differentiate for.
Also, adequately accounting for the dependencies between the bit allocation of different layers is not straight-forward. To meet these challenges, in this work we propose GradFreeBits: a novel joint optimization scheme for training dynamic QNNs, which alternates between gradient-based optimization for the weights, and gradient-free optimization for the bit allocation. Our method achieves better or on par performance with current state of the art low precision neural networks on CIFAR10/100 and ImageNet classification.
Furthermore, our approach can be extended to a variety of other applications involving neural networks used in conjunction with parameters which are difficult to optimize for.
\end{abstract}

\section{Introduction}
\label{intro}
In recent years, deep neural networks have been shown to be highly effective in solving many real world problems, creating an increased demand for their deployment in a variety of applications. However, such deep neural networks often require a large amount of computational resources for both training and inference purposes, which limits the adoption and spread of this technology in scenarios where large computational resources are not available. This issue is likely to persist in the near future, since recent trends have demonstrated that increased model size is often correlated with improved performance \cite{GPT3, tan2019efficientnet, BiT}. To mitigate this issue, recent efforts have been focused on developing specialized hardware to support the computational demands \cite{hardware_survey} as well as model compression methods in order to reduce them \cite{compression_survey}. These include broad families of techniques such as pruning \cite{pruning_survey}, knowledge distillation \cite{distillation_survey}, neural architecture search (NAS)\cite{nas_survey} and as is the focus of this paper, quantization \cite{hubara2017quantized}.

Quantization methods enable the computations performed by neural networks to be carried out with fixed point operations, rather than floating point arithmetic.
This improves the computational efficiency of neural networks and reduces their memory requirements. However, as with other compression methods, this typically comes at the cost of reduced performance \cite{compression_survey}. Recent efforts in the field have focused on improving the trade-offs between model compression and performance, by proposing a plethora of quantization schemes tailored for different scenarios.

\subsection{Quantization Methods}
Quantization schemes can first be divided into post training and quantization aware training schemes. Post training schemes decouple the tasks of model training and the quantization of its weights and/or activations. These methods are most suitable when the training data is no longer available when compressing the network, and only the trained network is available \cite{soudry1, postq1}. On the other hand, quantization aware training schemes perform both optimization tasks together, and do require training data, which tends to provide better performance.

In addition, quantization schemes can be further divided into methods which perform uniform or non-uniform quantization. Uniform quantization divides the real-valued domain into equally sized bins, whereas non-uniform methods do so with bins of varying sizes.  The former method tends to be more resource efficient and compatible with modern hardware accelerators, while the latter tends to provide better model performance \cite{li2019additive, jung2019learning}.

Quantization schemes can be further subdivided into per-channel methods, which utilize different quantization parameters for different channels within each layer, as opposed to per-layer methods, which use the same parameters for all channels in each layer, but different parameters for different layers. However, as with non-uniform quantization, per-channel methods can be difficult to implement efficiently on standard hardware components, making per-layer methods more feasible for deployment on edge devices.

For the reasons stated above, in this work we focus on per-layer, uniform quantization aware training with a different number of bits allocated per layer, referred to as dynamic quantization.
\subsection{Dynamic Quantization}\label{dynamicmotivation}
\begin{figure}
\vskip 0.1in
\begin{center}
\centerline{\includegraphics[width=\columnwidth]{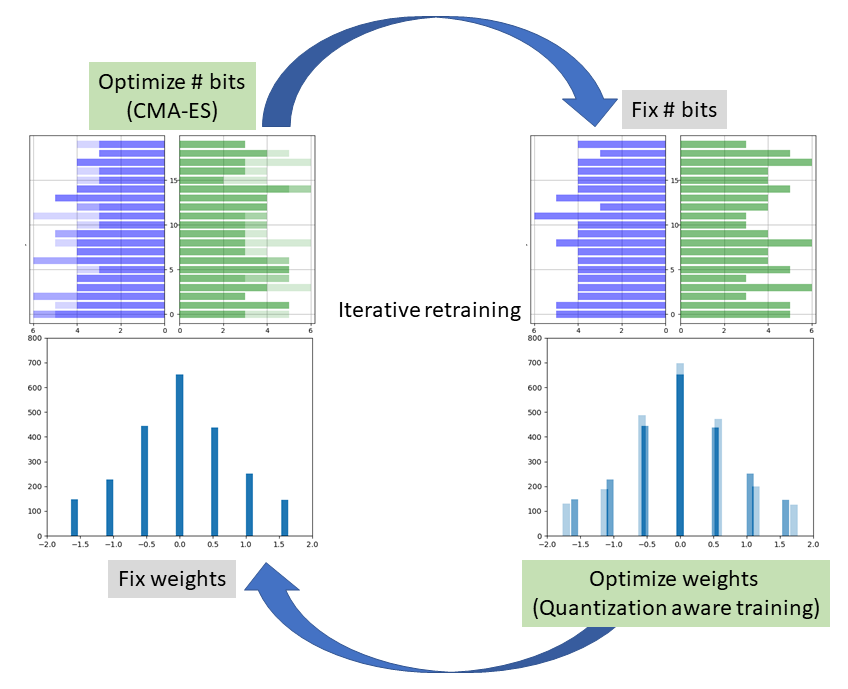}}
\caption{Our proposed training scheme, which uses iterative optimization of the model weights and bit allocation. Given a fixed allocation of bits for each layer, the weights are optimised using a quantization aware training procedure. Then, the bit allocation is optimized for those fixed weights and the process is repeated.}
\label{fig:OptScheme}
\end{center}
\vskip -0.1in
\end{figure}
Dynamic quantization aims to further improve the trade-offs between performance and computational requirements of QNNs for deployment on specialized edge devices. However, the proper allocation of bits between layers is combinatorial in nature and is hard to optimize. Recent efforts to tackle this problem include the use of reinforcement learning \cite{elthakeb2018releq, wang2019haq}, Hessian analysis \cite{dong2019hawqv1, dong2019hawqv2}, quantizer parametrization \cite{mpd}, and differentiable NAS approaches \cite{dnas_dynamic, li2020efficient, edmips}.  Among these methods, only the NAS approaches account for dependencies between bit allocations in different layers, by forming a super network that includes multiple branches for each precision at each layer. NAS approaches, however, are more expensive to train due to the multiple network branches. Here we offer an alternative approach to achieve the same goal.

\subsection{Our Contributions}
In this work we wish to optimize for the bit allocation of the network as a whole, and address the full interactions between different components of QNNs. To this end, we propose to utilize gradient-free optimization algorithms for the bit allocation. Such algorithms are known to perform well in difficult scenarios with complex dependencies between variables, while maintaining excellent sample efficiency during optimization \cite{conn2009introduction, rios2013derivative}. In particular, we use the algorithm Covariance Matrix Adaptation Evolution Strategy (CMA-ES) \cite{hansen2003reducing}, which aims to achieve the mentioned goals, though other alternatives may also be suitable for the task.

We propose a novel quantization aware training procedure for dynamic QNNs. In this procedure, CMA-ES is used for computing the optimal (per-layer) dynamic bit allocation for the weights and activations of the network. This method is interchangeably applied with gradient-based methods. That is, the network weights are updated by a gradient-based method, while the bit allocation is updated using the gradient-free method CMA-ES---see Fig. \ref{fig:OptScheme}. %illustrates our approach.

The advantages of our approach are as follows:
\begin{itemize}
\item Our training scheme for the dynamic bit allocation optimizes for the network as a whole. That is, it takes into account dependencies between layers of the neural network, as well as dependencies between weights and their bit allocation.
\item Our method for bit allocation is gradient-free, hence it can handle multiple (possibly non-differentiable) hardware constraints. This enables the QNN to be tailored to the resources of the specific edge devices it is to be deployed on.
% \item Our method requires less computational resources than the initial full-precision training.
\item The systematic combination of gradient-based and gradient-free optimization algorithms can be utilized in other applications and scenarios, e.g. systematic search of the network's other hyper-parameters.
\end{itemize}

\section{Preliminaries}
\subsection{Quantization Aware Training}\label{qat}
The quantization schemes of weights or activations in neural networks that we consider here can typically be broken into three steps: clip, scale and round. In the first step, the real values of the weights or activations are clipped to be within a range of values: i.e., $[-\alpha, \alpha]$ for symmetric signed quantization (typically weights) and $[0, \alpha]$ for unsigned quantization (typically activations, outputs of ReLU operations). Then, the range is mapped to the target integer range  $[-2^{b-1} + 1, 2^{b-1} - 1]$ and $[0, 2^{b} - 1]$, for signed and unsigned quantization respectively, where $b$ is the number of bits. The values $\alpha$ are  referred to as ``scales'' or clipping parameters and take on different values for different layers. The clipping parameters are optimized throughout the training procedure (along with the weights), so that they yield the lowest possible generalization error when quantization applied to the models' weights and/or activations.

Several approaches have been proposed in recent years, in order to optimize for the scales. In \cite{zhang2018lq} the scales are chosen based on quantization error minimization. \cite{choi2018pact} proposes differentiable clipping parameters which are learned through backpropagation together with the model's weights. This approach is further improved in \cite{li2019additive}, which also supports a computationally efficient \emph{non-uniform} quantization scheme. In this work we base our uniform quantization aware training scheme (for the weights and activations) on \cite{li2019additive}, while interchangeably utilizing CMA-ES for the dynamic bit allocation.

\begin{figure}
\vskip 0.1in
\begin{center}
\centerline{\includegraphics[width=7cm, height=3cm]{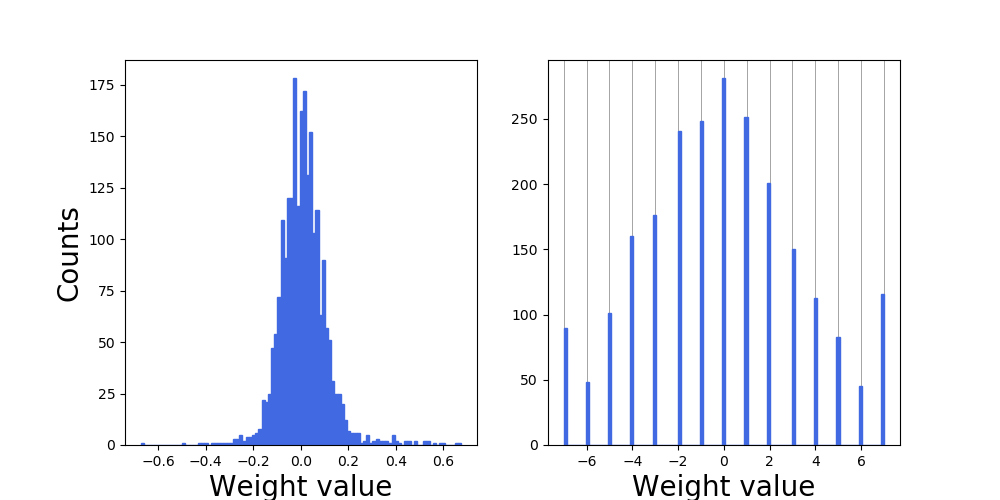}}
\caption{Example of a signed 4-bit quantized weights tensor and $\alpha=0.16$ (original-left, quantized-right). Values which are lesser or greater than the clipping parameters $\pm \alpha$ are clipped to $-\alpha$ or $+\alpha$ respectively, then scaled and rounded to the signed integer range. The value of 0 is preserved in this process.}
\label{fig:quantized_weights_example}
\end{center}
\vskip -0.1in
\end{figure}

To formally introduce the point-wise quantization operations we first define the quantization operator:
\begin{equation} \label{quantoperator}
    Q_b(x) = \frac{\mbox{round}((2^b - 1) \cdot x)}{2^b - 1},
\end{equation}
where $x$ is a real-valued tensor in [-1, 1] or [0, 1] for signed or unsigned quantization, respectively. $b$ is the number of bits that are used to represent $x$.
Given this operator, we use the reparameterized clipping function \cite{li2019additive} to define the quantized weights and activations.
\begin{eqnarray} \label{quantweights}
    W_b &=& \alpha_W Q_{b-1}(\mbox{clip}(\frac{W}{\alpha_W}, -1, 1))\\
    X_b &=& \alpha_X Q_{b}(\mbox{clip}(\frac{X}{\alpha_X}, 0, 1)) \label{quantacts}.
\end{eqnarray}
Here, $W, W_b$ are the real-valued and quantized weight tensors, $X, X_b$ are the real-valued and quantized input tensors, and $\alpha_W, \alpha_X$ are their associated scale (or clipping) parameters, respectively.
An example of Eq. \eqref{quantweights} being applied to a weights tensor, using 4-bit signed quantization (without multiplying by $\alpha_W$), can be seen in Figure \ref{fig:quantized_weights_example}.

We note that equations \eqref{quantoperator}-\eqref{quantacts} are used for training only. During inference, weights and activations are dynamically quantized, and all operations are performed using integers in mixed precision, while taking the scales into account.

In this quantization aware training scheme, both weights and activations are quantized during the forward pass, and during the backward pass, the Straight Through Estimator (STE) \cite{STE} is used to optimize the weights. That is, we ignore the derivative of $Q_b$ in the backward pass, and iterate on the floating point values of $W$ using the derivatives w.r.t $W_b$ in the SGD optimization. Given Eq. \eqref{quantweights}-\eqref{quantacts} the STE can also be used to calculate the gradients with respect to $\alpha_W, \alpha_X$ \cite{li2019additive}.
This enables the quantized network to be trained in an end-to-end manner using backpropogation.
To improve stability, \cite{li2019additive} also use weight normalization before quantization:
\begin{equation} \label{eq:normalize}
    \hat W = \frac{W - \mu}{\sigma  + \epsilon}.
\end{equation}
Here, $\mu$ and $\sigma$ are the mean and standard deviation of the weight tensor, respectively, and $\epsilon=10^{-6}$.

\subsection{CMA-ES}\label{prelim_cma}
Covariance Matrix Adaptation Evolution Strategies (CMA-ES) \cite{hansen2003reducing}, is a population based gradient-free optimization algorithm. It is known to be highly versatile and has been applied to a large variety of settings, such as RL \cite{heidrich2008evolution}, placement of wave energy converters \cite{neshat2019ahybrid}, hyper-parameter optimization \cite{loshchilov2016cma}, evolving game levels in Mario \cite{volz2018evolving}, adversarial attacks \cite{8917642} and more. It is designed to work in $d$-dimensional continuous spaces and optimize discontinuous, ill-conditioned and non-separable objective functions, in a black box optimization setting \cite{cma_bbob}.

At a high level, the optimization process of CMA-ES is as follows. At the $g$-th generation, a set of $\lambda$ $d$-dimensional samples are drawn from a multivariate normal distribution $\mathcal{N} (m^{(g)}, \mathcal{C}^{(g)})$:
\begin{equation}\label{cmaes}
    x_k^{(g+1)} \sim  m^{(g)} + \sigma^{(g)} \mathcal{N} (0, \mathcal{C}^{(g)}), \mbox{  for }k = 1,...,\lambda
\end{equation}
%Where $m^{(g)}$ is the mean of the population at generation $g$, $\mathcal{C}^{(g)}$ is the covariance matrix at generation $g$
Where $m^{(g)}$, $\mathcal{C}^{(g)}$ are the mean and covariance matrrix of the population at the previous generation, respectively. $\lambda$ is the population size and $\sigma^{(g)}$ is the step-size.

Once the samples are drawn from this distribution, they are evaluated and ranked based on the objective function.
Then, keeping only the top $\mu$ samples with the best objective values,  $m^{(g+1)}$, $\mathcal{C}^{(g+1)}$ and $\sigma^{(g+1)}$ are calculated using a set of update rules. Please refer to \cite{hansen2016cma} for more details.
These are then used to draw samples for the next generation using Eq. \eqref{cmaes}.
This process is repeated until one of several convergence criteria are full-filled, or until the algorithm exceeds its predefined budget of allowed samples.

Though CMA-ES is an effective gradient-free optimization algorithm, it has several downsides. First, its computational complexity is $O(d^2)$ in space and time \cite{hansen2003reducing}, where $d$ is the dimension of the parameter vector to be optimized.
Second, its convergence rate is often also $O(d^2)$ \cite{cma_bbob}, making it inefficient for solving high dimensional problems where $d$ is larger than a few thousands.

\section{The GradFreeBits Method}\label{method}
In this section we describe the components of our optimization procedure for dynamic QNNs.

\subsection{Motivation: CMA-ES for Dynamic Precision.}
We argue that CMA-ES is highly compatible with the problem of bit-allocation in QNNs. Here we assume the objective function is the differentiable loss function used during training, with additional possibly non-differentiable constraints related to computational requirements (exact details are provided below).

As recent evidence suggests \cite{wang2019haq, dong2019hawqv1}, the optimization landscape of the bit-allocation problem is likely to be discontinuous, ill-conditioned and amenable for optimization using gradient-free optimizers. Since the constraints may be non-differentiable, they can be sampled in a black-box setting, as is done in gradient-free optimization (CMA-ES) and reinforcement learning \cite{wang2019haq}.
Additionally, as shown in \cite{dong2019hawqv1, dong2019hawqv2}, the Hessian eigenvalues show large variations for different layers in QNNs, meaning that certain layers are typically more sensitive to changes in bit-allocation than others. This is in part what motivated us to choose CMA-ES for this work, as it is capable of adapting to high variations in the Hessian eigenvalues, and is therefore considered to be one of the best and widely used gradient-free methods.
Lastly, since neural networks typically have an order of tens to hundreds of layers, the problem of bit optimization falls well within the range of dimensionallity where CMA-ES is known to perform well, as discussed in the previous section.

\subsection{Setting the Stage for CMA-ES}\label{bitaloc}
To enable the gradient-free optimization algorithm to efficiently optimize the bit-allocation of a QNN, two items must be defined: its search space and the objective function.

\paragraph{Search Space}
We define the search space as a vector containing the bit allocation of the weights and activations in all the layers of the network, asides the first and the last layer, as these are quantized using a fixed allocation of 8 bits.
We found it beneficial to optimize the logarithm (base 2) of this vector, rather than the vector itself.
Thus, the vector to be optimized by CMA-ES is the log-precision vector:
\begin{equation}\label{cma_search_space}
    \mathbf{v} = [\mathbf{v_W}, \mathbf{v_X}] = \log_2 ([\mathbf{r_W}, \mathbf{r_X}]),
\end{equation}
where $\mathbf{r_W}, \mathbf{r_X}$ are the bit allocations of the network's weights and activations respectively, and $[\cdot, \cdot]$ is the concatenation operation.

\paragraph{Objective Function}
As mentioned in Section \ref{prelim_cma}, the objective function to minimize is a combination of the network's performance measure, the differentiable loss function, subject to a number of non-differrentiable computational constraints. More formally:
\begin{equation}
    \min_\mathbf{v} \textit{ } \mathcal{L}(\mathbf{v}; \mathbf{\theta}), \\
    \hspace{4pt} \mbox{ s.t. }  h_j(\mathbf{v}) \leq C_j \hspace{4pt} \mbox{ for } j=1,...,M
\end{equation}
Where $\mathcal{L}(\mathbf{v}; \mathbf{\theta})$ is the loss function over the training set, parameterized by network parameters $\mathbf{\theta}$, which are assumed to be fixed during the bit-allocation optimization stage.
Furthermore, $h_j(\mathbf{v})$ are the computational requirements for a given precision vector $\mathbf{v}$ (e.g., model size, inference time etc.), $C_j$'s are the target requirements that we wish to achieve, and $M$ is the number of constraints.
In our framework, we combine the constraints into the objective function using the penalty method \cite{penalty}:
\begin{equation}\label{objective}
    \min_\mathbf{v} \textit{ } \mathcal{L}(\mathbf{v}; \mathbf{\theta})
    + \sum_{j=1}^M \rho_j \max(0, h_j(\mathbf{v}) - C_j)^2,
\end{equation}
Where $\rho_j$ are balancing constraint parameters. This is similar to the approach taken in \cite{mpd}, but here it is applied to gradient-free optimization.

We define the computational constraints by comparing the requirements between the dynamic and static precision networks. For example, we wish that the model size will be equal to that of a static 4 bits allocation.
Specifically, we use constraints on the model size for the weights entries $\mathbf{v_W}$ and the mean bit allocation for the activation entries $\mathbf{v_X}$:
\begin{equation} \label{constraint1}
	    h_1({\mathbf{v}}) = MB({\mathbf{v_W}}), \hspace{4pt} C_1 = \beta_1 MB({\mathbf{v}^S_{\textbf{W}}})
\end{equation}
\begin{equation} \label{constraint2}
	    h_2({\mathbf{v}}) =
	    \frac{1}{L} \sum_{i=1}^L v_{X_i}, \hspace{4pt} C_2 =  \frac{\beta_2}{L} \sum_{i=1}^L v^S_{X_i}
\end{equation}
Where $MB(\cdot)$ calculates the model size given weight entries $\mathbf{v_W}$, $\mathbf{v}^S$ is the log-precision vector of the target static precision we wish to achieve, $L$ is the number of relevant layers, and $\beta_1, \beta_2>0$ control the target compression rates of the weights and activations, respectively.

These constraints are designed to limit the computational requirements, while  allowing the gradient-free optimization algorithm to explore non-trivial solutions which satisfy them.
It is important to note that though these constraints are only related to memory requirements, other constraints can easily be incorporated into our framework - such as power usage or inference time measurements, chip area, etc.

\subsection{Variance Reduction}\label{variance_reduction}
Variance reduction has been shown to improve the convergence rate of optimization algorithms \cite{variance_reduction}. The main source of variance in our objective function (Eq. \eqref{objective}) is in the first term, related to the performance of the model for different bit-allocations. There are two main causes of variance in this term: sub-sampling noise, caused by using small mini-batches of randomly selected samples, and sensitivity to quantization errors, which networks are typically not robust to.
To mitigate these issues and improve the convergence rate of CMA-ES we adopt two strategies for reducing the variance in the objective function, utilizing moving super-batches and quantization aware pretraining, which will be described promptly.

\begin{figure}
\vskip 0.1in
\begin{center}
\centerline{
\includegraphics[width=7.0cm, height=4.0cm]{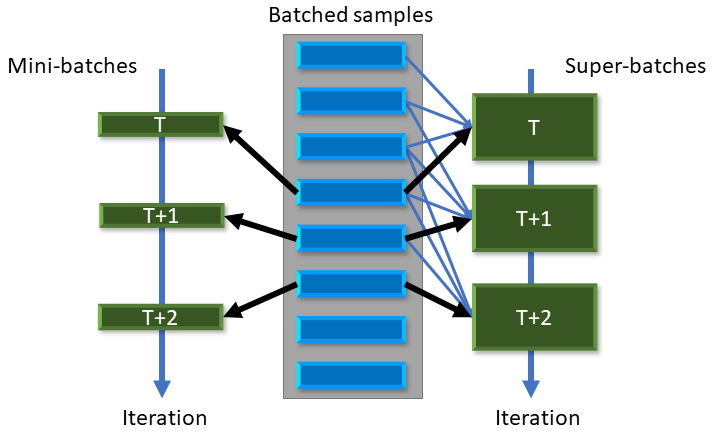}}
\caption{Batch replacement scheme used in moving super-batches, as compared to standard mini-batch replacement. At each iteration a single mini-batch is replaced, in a queue-like manner, creating a high overlap of samples between consecutive super-batches.}
\label{fig:moving_superbatch}
\end{center}
\vskip -0.1in
\end{figure}

\paragraph{Moving Super-batches}\label{moving_superbatch_section}
We define a moving super-batch as a set of mini-batches which are replaced in a queue-like manner. That is, in each iteration of the super-batch we replace part of the mini-batches within it. During each objective evaluation of CMA-ES, the entire super-batch is run through the model in order to calculate the first term of \eqref{objective}. The queue-like replacement scheme enables CMA-ES to encounter new data samples in each objective evaluation, but with a larger overlap of data samples as compared to standard SGD, where the entire mini-batch is re-sampled at each iteration. The process is illustrated in Figure \ref{fig:moving_superbatch}. Several strategies for the frequency of replacement can be considered, such as replacing one or more mini-batches after each objective evaluation, or doing so every fixed number of evaluations. These different settings are explored in the ablation study in Section \ref{ablation_study_section}.

\paragraph{Pretraining}\label{pretraining}
Another way to reduce variability in performance, is by reducing sensitivity to quantization errors. This can be done by pretraining the network, using the quantization aware training scheme, as described in section \ref{qat}. The bit allocation for this stage is static, meaning the weights and activations in all conv layers are quantized to the same target precision, except the first conv and last full-connected layers, which are always quantized to 8-bit precision.

\subsection{Gradient-free Steps}
%
%\tred{Benjy: The next sentence is also repeated in the following section, should we remove this?} \sout{Given a pretrained quantized network with a given bit allocation, (described above), the model is passed to the CMA-ES algorithm, in order to optimize}
We define gradient-free steps, as steps in which the CMA-ES algorithm optimizes the bit allocation (given the network's weights) according to the objective function in Eq. \eqref{objective}. In each step, multiple generations of samples of the log-precision vector $\mathbf{v}$ are evaluated on \eqref{objective}. At each objective evaluation, the sample of $\mathbf{v}$ is used to quantize the weights and activations of the model. Since CMA-ES typically operates in continuous spaces, the bit allocations of the weights and activations (which are positive integers) are extracted from $\mathbf{v}$ using the following formula:
\begin{equation} \label{extracting_precision}
    \mathbf{r_W} = \ceil{2^{\mathbf{v_W}}}, \mathbf{r_X} = \ceil{2^{\mathbf{v_X}}}
\end{equation}
Where $\ceil{x}$ is the "ceil" operator, which rounds its argument upwards to the nearest integer. Once the bit allocations have been inserted into the model, the loss \eqref{objective} is calculated over all the mini-batches in the super-batch, yielding the value of the objective for each of the sampled bit-allocations. This process gradually minimizes the value of the objective function and enables non-trivial bit-allocations to be found. We define each gradient-free step to include a predefined number of  objective evaluations. (this a hyper-parater, typically set to 512 evaluations). It is important to note that gradient-free steps require significantly less computational resources than traditional epochs, even if the number of mini-batches are matched. This is because they do not require backpropogation and the gradient-free optimizer computations are negligible compared to those performed by the model.
%\tred{W}e define a gradient-free step as a predefined number of  objective evaluations. (typically 512 evaluations).

%\tred{\textbf{Eran: maybe it's a good place to brag about the grad-free steps being much less expensive than grad based. that is, (1) each step may have less evaluations than an epoch (at least the setup here). (2) our evaluations do not require a backward pass, hence less expensive.}  }

\subsection{Iterative Alternating Retraining}
The primary innovation in this work is the combination of gradient-based training sessions of the model weights and gradient-free training sessions of the bit-allocation, which we refer to as iterative alternating retraining.

After the pretraining stage, the model is passed to the gradient-free optimizer CMA-ES to optimize its bit allocation for a session of a predefined number of steps $N_{GF}$. This adapts the bit allocation to the model weights, which are fixed at this stage in their floating point values. This maximizes the performance of quantized networks, subject to the computational constraints (Eq. \eqref{objective}), using the optimization process as described in section \ref{prelim_cma}.

Once the gradient-free session is completed, the best bit allocation found by CMA-ES is passed to the gradient-based optimizer, for a session of a predefined number of epochs $N_{GB}$. This adapts the model weights to the bit allocation, which is kept fixed, using the quantization aware training scheme described in Section \ref{qat}. Once this session is completed, the model is returned to the gradient-free optimizer to further improve the bit allocation, given the new model weights. This cycle is then repeated a number of times, or until the performance and computational requirements are satisfactory. The process is illustrated in Fig. \ref{fig:OptScheme}.

To increase stability in the handover process between the optimization algorithms, we found it beneficial to restart CMA-ES after each gradient-based session and handover only the best model weights and bit allocation (with lowest loss/objective values) found across all sessions, as opposed to handing over the best versions found in each session.

\section{Experiments}

\begin{figure}
\vskip 0.1in
\begin{center}
\centerline{
\includegraphics[width=\columnwidth, height=3.8cm]{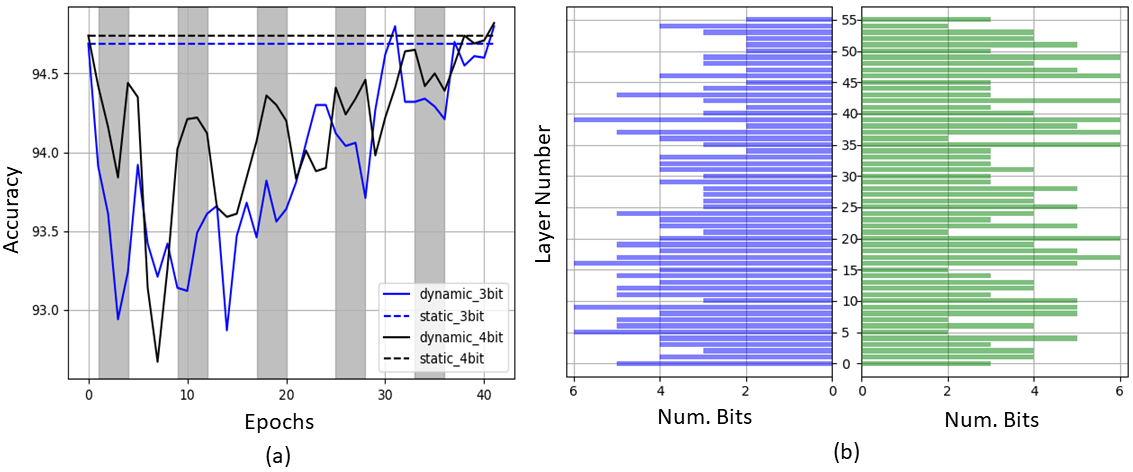}
}
\caption{(a) Convergence plot for a 3 and 4-bit dynamically quantized ResNet56 on CIFAR10, during the retraining process. Grey regions correspond to gradient-free steps, while white regions correspond to gradient-based epochs. (b) The bit allocations for the weights (blue, left) and activations (green, right) of the 4-bit ResNet56.}
\label{fig:res56_cifar10}
\end{center}
\vskip -0.1in
\end{figure}

To quantitatively compare our dynamic quantization scheme (GradFreeBits) to other related works, we apply it to several neural network architectures for image classification tasks. Throughout all the experiments, we consider similar numbers of bits (on average, in the dynamic case) for the weights and activations. Since we use the ReLU activation in all the settings, the activations are quantized using unsigned integers. The weights are quantized symmetrically using signed integers, so for example, 2-bit quantization will result in ternary weights in $\{-1,0,1\}$.

Furthermore, in this section we also perform an ablation study, to identify the effects that different settings of the proposed system have on the performance metrics of the resulting dynamically quantized models.

We compare our approach to the following related works that use uniform quantization, with either static (S) or dynamic (D) bit-allocation schemes:
LQ-Nets(S) \cite{zhang2018lq}, PACT(S) \cite{choi2018pact}, APoT(S) \cite{li2019additive}, DSQ(S) \cite{gong2019differentiable}, BSGD(S) \cite{bcgd}, TQT(D) \cite{tqt},  MPD(D) \cite{mpd}, EBS(D)\cite{li2020efficient}, EDMIPS(D) \cite{edmips}, HAQ(D) \cite{wang2019haq}, HAWQ-V1(D) \cite{dong2019hawqv1}, HAWQ-V2(D) \cite{dong2019hawqv2}, WNQ(S) \cite{wnq}, RES(S) \cite{res}, DoReFa-Net(S) \cite{drfn}. Lastly, we denote our our proposed method GradFreeBits as GFB(D).

\subsection{CIFAR 10/100} \label{cifar10_section}
The CIFAR10 and CIFAR100 image classification benchmarks \cite{cifar10} have 10 and 100 classes respectively, both containing 50k train and 10k test 32x32 RGB images. For these datasets, pretraining \textit{from scratch} was conducted for 400 epochs, using a mini-batch size of 128, SGD optimizer with momentum of 0.9, cosine learning rate decay, with a maximum learning rate of 0.1 and warm-up of 10 epochs. Data augmentation used during the pretraining and iterative retraining stages included: random horizontal flips and crops, and mixup \cite{mixup}. To approximately satisfy the constraints of \eqref{objective}, we chose $\beta_1=\beta_2=0.7$ and $\rho_1=\rho_2=0.5$. This allows the penalty constraints to be slightly violated, while keeping model size and mean precision below their actual targets. We also limit the search space, such that all values of $\textbf{v}$ are in $[0, 3]$.

Given the pretrained model, we applied 5/3 rounds of the iterative alternating retraining algorithm for CIFAR10/100 respectively. In each round, we applied 4 steps in the gradient-free session and 4/16 epochs in the gradient-based session for CIFAR10/100 respectively. At each gradient-free step of CMA-ES we applied 512 objective evaluations on super-batches, each containing 32 mini-batches of 128 data samples. After each of the 512 objective evaluations we swap only one mini-batch in the super-batch as illustrated in Fig. \ref{moving_superbatch_section}.
%For CIFAR10/100, the entire procedure requires objective evaluations that are comparable to \sout{440/460} \tred{1260/953} epochs respectively (400 epochs for pretraining the quantized networks from scratch, 20/12 gradient-free steps, and 20/48 gradient-based epochs).

%the iterative alternating retraining

%using 32 mini-batches of 128 data samples in each super-batch and grad-free epochs of 512 \sout{generations} \tred{objective evaluations}. We performed 3 \sout{alternating minimization} \tred{iterative alternating retraining} rounds, where in each we apply 4 epochs in each gradient-free session and 8 epochs in each gradient-based session.

%After each \sout{generation} \tred{objective evaluation} we swap only one mini-batch in the super-batch.
%This amounts to 9.5 GPU-hours for Resnet20 and 16 GPU-hours for Resnet56, on a single GeForce RTX 2080 Ti.}

\begin{table}[t]
\caption{Top1 accuracy of quantized network on CIFAR10/100. (S) and (D) denote static and dynamic quantization, while \textbf{*} denotes experiments which use 2-bit weights and 4-bit activations.}
\label{tab:cifar10}
\vskip 0.15in
\begin{center}
\begin{small}
\begin{sc}
\begin{tabular}{lccr}
\toprule
Model & Method  & 3/3 & 4/4 \\
\midrule

Res.20    & PACT(S)          & 91.1 & 91.7  \\
CIFAR10    & LQ-Nets(S)        & 91.6       & -  \\
FP 93.3  & BCGD(S)           & 91.2 * & 92.0 \\
          & TQT(D)           & 90.4 * & - \\
          & MPD(D)           & 91.4 * & - \\
          & EBS(D)             & 92.7 & 92.9 \\
          \rowcolor{gray!20}
          & GFB(D) (ours)        & \textbf{93.3} & \textbf{93.6}  \\
           \rowcolor{gray!20}
         & GFB(D) (ours)        & \textbf{93.4}* & -  \\
\hline
\hline

Res.56 CIFAR10 & EBS(D)      & 94.1 & 94.3 \\
\rowcolor{gray!20}
FP 95.1           & GFB(D) (ours)  & \textbf{94.8} & \textbf{94.9} \\
\hline
\hline

Res.20    & DoReFa-Net(S)       & 68.4 & 68.9  \\
CIFAR100   & Res(S)        & 68.3       & 68.7  \\
FP 70.35   & LQ-Nets(S)        & 68.4       & 69.0  \\
          & WNQ(S)        & 68.8       & 69.0  \\
          \rowcolor{gray!20}
          & GFB(D) (ours)   & \textbf{69.6} & \textbf{70.6}  \\
\bottomrule
\end{tabular}
\end{sc}
\end{small}
\end{center}
\vskip -0.1in
\end{table}

The results for the CIFAR10 dataset are presented in Table \ref{tab:cifar10}. Our method out-performs the previous state of the art EBS(D) for both ResNet20 and ResNet56 models, at both dynamic precision settings: e.g. +0.7\% for 4-bit ResNet20 and +0.6\% for 4-bit ResNet56. Table \ref{tab:cifar10} also includes the results for the  CIFAR100 dataset. For this benchmark, our method also outperforms all other related works: by 1.6 for 4-bit ResNet20 and 0.8 for fore 3-bit ResNet20. %Though the trend of decreasing performance with decreasing precision was observed in the CIFAR10 benchmark, it is enhanced in CIFAR100, likely due to the increased difficulty of the task.

%\sout{Though dynamic quantization improves the performance of the Resnet56 model, the performance of the dynamically quantized Resnet20 model remains unchanged. One possible explanation for this is the expressiveness of Resnet20 model is already saturated for this benchmark, making it difficult for the bit allocation to improve its performance. }

\subsection{ImageNet}\label{imagenet_section}
The ImageNet \cite{imagenet} image classification benchmark has 1K classes, 1.2M train and 150K test RGB images. For all models, we used pretrained weights from TorchVision \cite{torchvision}.
Additional quantization aware pretraining was then conducted for 30 epochs, using mini-batches of size 128, SGD optimizer with momentum of 0.9, and a cosine learning rate decay with a maximum of 0.001 and a warm-up of 3 epochs.

Data augmentations are identical to those used in the CIFAR10/100 experiments (above), with the addition of image resize to $256 \times 256$ and random crops to $224 \times 224$. These are used for both pretraining and iterative alternating retraining stages. Furthermore, $\beta_1=\beta_2=0.7$, $\rho_1=\rho_2=0.5$ are used to satisfy the constraints and the search space for $\textbf{v}$ is limited to $[0, 3]$. Three rounds of iterative alternating retraining were applied. In each round, we applied 4 steps in the gradient-free session and 4 epochs in the gradient-based session.
In each gradient-free step of CMA-ES we applied 512 objective evaluations on super-batches, each containing 8 mini-batches of 128 data samples, swapping one-minibatch in the super-batch after each objective evaluation. %\sout{with 8 mini-batches of 128 data samples in each super-batch, grad-free epochs of 512 objective evaluations, swapping one-minibatch after each objective evaluation, 4 epochs for each gradient-free session and 4 epochs for each gradient-based session, with a total of 3 sessions each being applied}.
The entire procedure requires objective evaluations that are comparable to 47 epochs (30 gradient-based pretraining epochs, 12 gradient-free steps, and 12 gradient-based epochs).
%We note that the gradient-free steps are less computationally intensive than the gradient-based epochs, since they do not require backpropagation.
%This amounts to 5 GPU-days for ResNet18 and 8 GPU-days for ResNet50, on a single GeForce RTX 2080 Ti. \textbf{Eran: this is quite long here for 54 epochs. I'm not sure it's really 54. Let's talk about it.}}
%At each round we applied 4 steps in each gradient-free session and 4\tred{/16} epochs in each gradient-based session \tred{for CIFAR10/100 respectively}.
\begin{table}[t]
\caption{Top1 accuracy of quantized networks on ImageNet. (S) and (D) denote static and dynamic quantization, ($\cdot$) denotes model size, measured in MB. $\approx$ is used when only graphical results were available for comparison. \textbf{*} denotes experiments which use 2-bit weights and 4-bit activations.}
\label{tab:ImageNet}
\vskip 0.15in
\begin{center}
\begin{small}
\begin{sc}
\resizebox{\columnwidth}{!}{%
\begin{tabular}{lccr}
\toprule
Model & Method & 3/3 & 4/4 \\
\midrule
Res.18         & BCGD(S)           & -  & 67.4(-) \\
FP 70.7    & LQ-Nets(S)         & 68.2(6.1)            & 69.3(7.4) \\
(46.8)   & DSQ(S)          & 68.7(6.1)            & 69.6(7.4) \\
          & PACT(S)       & 68.1(6.1)            & 69.2(7.4)  \\
          & APoT(S)       & 69.4(4.6)            & - \\
          & EBS(D)         & 69.5(-)     & 70.2(-) \\
          & EDMIPS(D)      & $\approx$67(-)       & $\approx$68(-)  \\
          & TQT(D)         &     -     & 69.5(5.6)  \\
          & MPD(D)         & -      & 70.1(\textbf{5.4})  \\
         \rowcolor{gray!20}
         % & GFB(D) (ours)     & \textbf{69.6}(\textbf{4.9})   & \textbf{70.3}(\textbf{5.4})  \\
          & GFB(D) (ours)          & 69.2(\textbf{3.6})   & \textbf{70.3}(\textbf{5.4})  \\

\hline \hline
Res.50    & DoReFa-Net(S)      & 69.9(16.6)            & 71.4(19.4) \\
FP 76.4   & LQ-Nets(S)      & 74.2(16.6)            & 75.1(19.4) \\
(97.5)    & PACT(S)       & 75.3(16.6)            & \textbf{76.5}(19.4)  \\
          & EDMIPS(D)      & $\approx$72.5(-)     & $\approx$74.0(-)   \\
          & HAQ(D)       & 75.4(9.2)        & 76.1(\textbf{12.1}) \\
          & HAWQ-V1(D)      & 75.5(\textbf{8.0})  \textbf{*}           & -  \\
          & HAWQ-V2(D)     & \textbf{75.8}(\textbf{8.0})  \textbf{*}        & - \\
         \rowcolor{gray!20}
          & GFB(D) (ours)     & 75.7(9.6) & 76.1(12.8)  \\
\bottomrule
\end{tabular}}
\end{sc}
\end{small}
\end{center}
\vskip -0.1in
\end{table}

% \noindent The results for the Imagenet data are presented in table \ref{ImageNet}. For the ResNet18 model, our method slightly outperforms the previous state of the art EBS(D) \cite{li2020efficient} for 3 and 4-bits dynamic precision $+0.1\%$. Furthermore, both dynamically quantized models achieve the smallest model size in their category.

\noindent The results for the ImageNet dataset are presented in Table \ref{tab:ImageNet}. For the Renet18 model, our method slightly outperforms the previous state of the art EBW(D)\cite{li2020efficient} for 4-bits dynamic precision $+0.1\%$, though slightly under-performs for 3-bits $-0.3\%$. However, both dynamically quantized models achieve the smallest model size in their category.

%\sout{For the 4-bit resnet50 models, our method is on par with the previous state of the art HAQ(D)} \cite{wang2019haq}, \sout{even though it uses non-uniform quantization. Our 3-bit ResNet50 achieves similar Top1 accuracy to that of HAWQ2(D)} \cite{dong2019hawqv2} , \sout{though with a larger model size $+1.6MB$.}
For the 3-bit weights and 3-bit ResNet50, our method achieves similar Top1 accuracy to that of HAWQ2(D) \cite{dong2019hawqv2} with 2-bit weights and 4-bit activations, though with a larger model size $+1.6MB$. However, our 4-bit ResNet50 falls behind PACT(S) \cite{choi2018pact} in performance $-0.4\%$, though with a significantly smaller model size $-6.6MB$. This is on par with the previous state of the art method HAQ(D) \cite{wang2019haq}, even though it uses \emph{non-uniform}, quantization which can achieve a higher accuracy than uniform.

For these experiments we can observe the advantage of the dynamically quantized models compared to the statically quantized ones. Even though the dynamically quantized models require less memory, they tend to obtain higher or comparable Top1 accuracy scores.

\subsection{Ablation Study}\label{ablation_study_section}
In this section we examine the effects that different system settings have on the performance of our dynamically quantized models. More specifically, we examine the effects of pretraining, iterative alternating retraining, the different settings of moving super-batches and the number of mini-batches they contain. To enable efficient comparison, all experiments are conducted with 4-bit dynamically quantized ResNet20 models on CIFAR100 with the same hyperparameters used in \ref{cifar10_section}. (learning rate, number of gradient-based/free epochs, etc.).

We consider the following settings for replacing mini-batches in the super-batch with new batches of randomly selected training samples: \textbf{NR} - no mini-batches are replaced in the super-batch, \textbf{EB} - single mini-batch replacement after each gradient-free step, \textbf{EF} - replacement of all mini-batches after each gradient-free step, \textbf{SB} -  single mini-batch replacement after each objective evaluation, \textbf{SF} - replacement of all mini-batches after each objective evaluation. The test without iterative alternating retraining runs all gradient-free steps first, followed by all the gradient-based epochs. The test without pretraining, uses the full-precision model as the initial weights.
%
% \begin{table}[t]
% \caption{Ablation study of a 4-bit dynamically quantized ResNet20 on CIFAR100, for various system settings. We use the shorthand: "SS." for super-batch setting, "NMB." for number of mini-batches in the super-batch, "IT.-RET." for iterative alternating retraining, "PRET." for pretraining, "ACC." for accuracy and "SZ." for model size.(4-bit static size is 0.148MB)}
% \label{ablation_system_settings}
% \vskip 0.15in
% \begin{center}
% \begin{small}
% \begin{sc}
% \begin{tabular}{lccccr}
% \toprule
% ss. & nmb. & it.-ret. & pret. & Acc. & Sz.(MB) \\
% \midrule
% SB      & 32 & $\surd$ & $\surd$ & \textbf{70.61}          & 0.164 \\
% \hline
% EB      & 32 & $\surd$ & $\times$& 66.99          & 0.158 \\
% EB      & 32 & $\times$& $\surd$ & 70.43          & 0.183 \\
% \hline
% NR      & 32 & $\surd$ & $\surd$ & 70.21          & 0.161      \\
% EB      & 32 & $\surd$ & $\surd$ & 70.26          & 0.166      \\
% EF      & 32 & $\surd$ & $\surd$ & 70.26          & 0.161 \\
% SF      & 32 & $\surd$ & $\surd$ & 70.25          & 0.168 \\
% \hline
% SB      & 4 & $\surd$ & $\surd$ & 69.68          & 0.157  \\
% SB      & 8 & $\surd$ & $\surd$ & 70.25          & 0.185 \\
% SB      & 16 & $\surd$ & $\surd$ & 70.18          & \textbf{0.152} \\
% SB      & 64 & $\surd$ & $\surd$ & 70.19          & 0.154 \\
% \bottomrule
% \end{tabular}
% \end{sc}
% \end{small}
% \end{center}
% \vskip -0.1in
% \end{table}

\begin{table}[t]
\caption{Ablation study of a 4-bit dynamically quantized ResNet20 on CIFAR100, for various system settings. We use the shorthand: "SS." for super-batch setting, "NMB." for number of mini-batches in the super-batch, "IT.-RET." for iterative alternating retraining, "PRET." for pretraining and "ACC." for accuracy.}
\label{ablation_system_settings}
\vskip 0.15in
\begin{center}
\begin{small}
\begin{sc}
\begin{tabular}{lccccr}
\toprule
variable & ss. & nmb. & it.-ret. & pret. & Acc. \\
\midrule
Baseline & SB      & 32 & $\surd$ & $\surd$ & \textbf{70.61} \\
\hline
Components & SB      & 32 & $\surd$ & $\times$& 66.99 \\
 & SB      & 32 & $\times$& $\surd$ & 70.43 \\
\hline
Super-batch & NR      & 32 & $\surd$ & $\surd$ & 70.21  \\
Setting & EB      & 32 & $\surd$ & $\surd$ & 70.36  \\
 & EF      & 32 & $\surd$ & $\surd$ & 70.26   \\
 & SF      & 32 & $\surd$ & $\surd$ & 70.39   \\
\hline
Super-batch & SB      & 4 & $\surd$ & $\surd$ & 69.68    \\
Size & SB      & 8 & $\surd$ & $\surd$ & 70.25    \\
 & SB      & 16 & $\surd$ & $\surd$ & 70.18   \\
& SB      & 64 & $\surd$ & $\surd$ & 70.26   \\
\bottomrule
\end{tabular}
\end{sc}
\end{small}
\end{center}
\vskip -0.1in
\end{table}

\noindent The results of the ablations study are presented in Table \ref{ablation_system_settings}. The $-3.6\%$ accuracy degradation demonstrates that pretraining plays a crucial role in reducing performance degradation due to changes in bit-allocation. It also seems that iterative alternating retraining, as opposed to separating the bit-optimization and weight-optimization stages, leads to a small $+0.2\%$ increase in performance, demonstrating the added value of this approach. Regarding the super-batch settings, it seems that the optimal setting is to use 32 mini-batches and replace a single batch after each objective evaluation (SB), also leading to a performance increase of $+0.2\%$. For these reasons, we used these settings used in the rest of the experiments.

\section{Discussion}
%
%
%General trends we've observed.
%Limitations of the approach.
%Special considerations taken for hyperparameter settings.
%Lessons we've learned.
%
Though the proposed method achieves favorable trade-offs between performance and model size, several items need be taken into account when applying it to new problems.

First, since CMA-ES is designed to operate in continuous search spaces and the search space considered here is discrete, rounding operations are required to ensure compatibility (Eq. \eqref{extracting_precision}). This creates plateaus of constant values in the objective function, which may harm the convergence rate. To tackle this, we may use other gradient-free algorithms, which are more suitable for discrete optimization, and may lead to even better results using our framework.

Secondly, though this work shows that the CMA-ES optimizer is highly effective in a variety of QNNs, applying it to larger networks (such as EfficientNet B7 \cite{tan2019efficientnet}) may require more generations and objective evaluations in order to achieve the comparable improvements in the performance and compression rate. This is because the convergence rate of CMA-ES is typically $O(L^2)$ (see section (\ref{prelim_cma})), where $L$ is the number of layers.
That said, if these consideration are taken into account, the combination of gradient-based and gradient-free optimization methods can be applied to achieve highly competitive results.

\section{Conclusion}
We proposed GradFreeBits, a novel framework for optimizing the bit-allocation in dynamically quantized neural networks, which enables neural networks to be customized to meet multiple hardware constraints. The framework is based on the combination of a gradient-based quantization aware training scheme for the weights and gradient-free optimization of the bit-allocation, based on CMA-ES. By benchmarking our method on multiple image classification tasks, we find that our method often outperforms several related dynamic and static quantization methods, in both accuracy and model size. We believe our method will help accelerate the deployment of low-precision neural network on resource constrained edge-devices, by enabling the network compression to be tailored to the specific hardware requirements. Additionally, the proposed framework for combining gradient-free and gradient-based optimization in an iterative alternating retraining scheme is quite general, making it likely easy to apply to other applications.

Future work in this direction includes utilizing additional constraints, such as FLOPS count and measurements from hardware simulators (such as BISMO \cite{bismo}) - including inference time, power consumption and chip area. Furthermore, it is important to evaluate this method on new tasks such as object detection and semantic segmentation, as these tasks are more difficult than image classification, perhaps making the models more sensitive to changes in bit allocation.
We hope this framework will provide a foundation for future works related to the combination of gradient-free and gradient-based neural network training, and are looking forward to more contributions in this direction.

\small
\bibliography{Qbib.bib}
\bibliographystyle{icml2021}

\newpage

\normalsize

\section{Appendix: CMA-ES}\label{prelim_cma}

In this section we provide a very brief overview of the main derivations that are used in CMA-ES, without including any theoretical background for the structure of these update rules, or the choices of hyperparameters. For more details regarding these topics, we refer the curious reader to \cite{hansen2016cma}---we follows the same notation as this paper here. 

Covariance Matrix Adaptation Evolution Strategies (CMA-ES) \cite{hansen2003reducing}, is a population based gradient-free optimization algorithm.
At a high level, the optimization process of CMA-ES is as follows. At the $g$-th generation, a set of $\lambda$ $d$-dimensional samples 
$\boldsymbol{x}_{k}  \in \mathbb{R}^d$
are drawn from a multivariate normal distribution $\mathcal{N} (m^{(g)}, \mathcal{C}^{(g)})$:
\begin{equation}\label{cmaes2}
    \boldsymbol{x}_{k}^{(g+1)} \sim \boldsymbol{m}^{(g)}+\sigma^{(g)} \mathcal{N}\left(\mathbf{0}, \boldsymbol{C}^{(g)}\right), \mbox{  for }k = 1,...,\lambda
\end{equation}
%Where $m^{(g)}$ is the mean of the population at generation $g$, $\mathcal{C}^{(g)}$ is the covariance matrix at generation $g$
Where $\boldsymbol{m}^{(g)}$, $\boldsymbol{C}^{(g)}$ are the mean and covariance matrix of the population at the previous generation, respectively. $\lambda$ is the population size and $\sigma^{(g)}$ is the step-size. 

Once the samples are drawn from this distribution, they are evaluated and ranked based on their objective function values. These ranked samples $\boldsymbol{x}_{i: \lambda}^{(g+1)}$ are used to calculate $m^{(g+1)}$, $\mathcal{C}^{(g+1)}$ and $\sigma^{(g+1)}$ of the next generation, using a set of update rules which are provided below.

\subsection{Hyperparameters}
% CMA-ES relies on a several hyperparameters in order to perform optimization. (referred to as "parameters" in  \cite{hansen2016cma}).
% Mainly, $d_{\sigma}$ is a damping parameter and $c_{1}, c_{\mu}, c_{\mathrm{c}}, c_{\mathrm{m}}, c_{\sigma}$ are "momentum"-like parameters, which control the amount of information retained from previous updates.

% An important choice of hyparamerameters are the "recombination weights" $w_i$, which play a vital role in the update rules. Typically they are chosen such that:

CMA-ES uses several hyperparameters in order to perform optimization \cite{hansen2016cma}. These include a dampling parameter $d_{\sigma}$ and $c_{1}, c_{\mu}, c_{\mathrm{c}}, c_{\mathrm{m}}, c_{\sigma}$ which are "momentum"-like parameters, which control the amount of information retained from previous generations. Furthremore, $w_i$ are known as the "recombination weights", which are used in most update rules. They are typically chosen such that

\begin{equation}\label{cmaes}
\sum_{j=1}^{\lambda} w_{j} \approx 0, \quad \sum_{i=1}^{\mu} w_{i}=1, \quad w_{1} \geq \cdots \geq w_{\mu}>0.
\end{equation}

These are also used to calculate the effective population size for recombination: $\mu_{\mathrm{eff}}=\left(\sum_{i=1}^{\mu} w_{i}^{2}\right)^{-1}$. For more details regarding the specific choices of these hyparamerameters, please refer to \cite{hansen2016cma}.

\subsection{Mean Update Rule}
% The first update rule is for updating the mean $\boldsymbol{m}^{(g+1)}$ is
As mentioned above, several update rules are employed in CMA-ES. The first of those is the update of the mean $\boldsymbol{m}^{(g+1)}$:
\begin{equation}\label{cmaes}
    \boldsymbol{m}^{(g+1)}=\boldsymbol{m}^{(g)}+c_{\mathrm{m}} \sum_{i=1}^{\mu} w_{i}\left(\boldsymbol{x}_{i: \lambda}^{(g+1)}-\boldsymbol{m}^{(g)}\right).
\end{equation}

\subsection{Covariance Matrix Update Rule}
The covariance matrix requires auxiliary vectors in order to construct its update rule, these are calculated using:

\begin{eqnarray} \label{auxilary_vectors}
    \boldsymbol{p}_{\mathrm{c}}^{(g+1)} &=& \left(1-c_{\mathrm{c}}\right) \boldsymbol{p}_{\mathrm{c}}^{(g)} + \tilde c_{\mathrm{c}} \frac{\boldsymbol{m}^{(g+1)}-\boldsymbol{m}^{(g)}}{\sigma^{(g)}}\\
    \boldsymbol{y}_{i: \lambda}^{(g+1)} &=& \left(\boldsymbol{x}_{i: \lambda}^{(g+1)}-\boldsymbol{m}^{(g)}\right) / \sigma^{(g)},
\end{eqnarray}
where $\tilde c_{\mathrm{c}} = \sqrt{c_{\mathrm{c}}\left(2-c_{\mathrm{c}}\right) \mu_{\mathrm{eff}}}$. These auxiliary vectors are then used to construct the rank-$\mu$ and rank-$1$ update matrices:
\begin{eqnarray} \label{auxilary_matricies}
    \boldsymbol{C}_{\mu}^{(g+1)} &=&  \sum_{i=1}^{\lambda} w_{i} \boldsymbol{y}_{i: \lambda}^{(g+1)} \left(\boldsymbol{y}_{i: \lambda}^{(g+1)}\right)^{\boldsymbol{\top}}\\
    \boldsymbol{C}_{1}^{(g+1)} &=&  \boldsymbol{p}_{\mathrm{c}}^{(g+1)} \boldsymbol{p}_{\mathrm{c}}^{(g+1)^{\top}}.
\end{eqnarray}
Finally, by defining $c_{old} = \left(1-c_{1}-c_{\mu} \sum_{i=1}^{\lambda} w_{i}\right)$, we get the update rule for the covariance matrix:
\begin{equation}\label{covariance_matrix_update}
\boldsymbol{C}^{(g+1)} =  c_{old} \boldsymbol{C}^{(g)} + c_1 \boldsymbol{C}_{1}^{(g)} + c_{\mu} \boldsymbol{C}_{\mu}^{(g+1)}.
\end{equation}

\subsection{Step Size Update Rule}
The last update rule is for the step size, which also requires use of an auxiliary vector: 
\begin{equation}
\boldsymbol{p}_{\sigma}^{(g+1)}=\left(1-c_{\sigma}\right) \boldsymbol{p}_{\sigma}^{(g)}+\tilde c_{\sigma}  \boldsymbol{C}^{(g)^{-\frac{1}{2}}} \frac{\boldsymbol{m}^{(g+1)}-\boldsymbol{m}^{(g)}}{\sigma^{(g)}},
\end{equation}
Where $\tilde c_{\sigma} = \sqrt{c_{\sigma}\left(2-c_{\sigma}\right) \mu_{\mathrm{eff}}}$. This is then used to construct the last update rule, which is for the step size:
\begin{equation}
\sigma^{(g+1)}=\sigma^{(g)} \exp \left(\frac{c_{\sigma}}{d_{\sigma}}\left(\frac{\left\|\boldsymbol{p}_{\sigma}^{(g+1)}\right\|}{E\|\mathcal{N}(\mathbf{0}, \mathbf{I})\|}-1\right)\right),
\end{equation}
where ${E\|\mathcal{N}(\mathbf{0}, \mathbf{I})\|} \approx \sqrt{n}+\mathcal{O}(1 / n)$ is the expectation of the $l_2$ norm of samples drawn from $\mathcal{N}(\mathbf{0}, \mathbf{I})$.

\subsection{Next Generation}
Once  $\boldsymbol{m}^{(g+1)}$, $\boldsymbol{C}^{(g+1)}$ and $\sigma^{(g+1)}$ have been calculated, they are inserted into \eqref{cmaes2}, so that the next generation of samples can be drawn and the process repeated, until the convergence criteria are full-filled.

%\end{document}

\end{document}